\documentclass[sigconf]{acmart}
\AtBeginDocument{%
  }

\setcopyright{acmlicensed}
\copyrightyear{2027}
\acmYear{2027}
\acmDOI{XXXXXXX.XXXXXXX}

  \acmConference[SIGCSE TS 2027]{the 57th ACM Technical Symposium on Computer Science Education}{February 17--20, 2027}{Sacramento, CA, USA}
\acmISBN{978-1-4503-XXXX-X/2018/06}

\usepackage{booktabs}
\usepackage{hyphenat}
\begin{document}

\title{An Educator-Guided LLM Pedagogical Agent for Scaffolded Feedback in Conceptual Database Design}

\author{Sara Riazi}
\affiliation{%
  \institution{University of Illinois Chicago}
  \city{Chicago, IL}
  \country{USA}
}
\email{riazi@uic.edu}

\author{Pedram Rooshenas}
\affiliation{%
  \institution{University of Illinois Chicago}
  \city{Chicago, IL}
  \country{USA}
}
\email{pedram@uic.edu}

\renewcommand{\shortauthors}{Riazi and Rooshenas}

\begin{abstract}
We present an educator-guided LLM pedagogical agent for scaffolded
feedback in conceptual database design. Integrated into an
entity--relationship diagram (ERD) editor, the system grounds feedback in the
student artifact, assignment requirements, educator-authored rubrics,
and instructional resources. Its architecture separates hidden,
artifact-grounded diagnosis from the workflow that controls the form
and disclosure level of student-facing support.

We instantiate the architecture as a four-stage workflow progressing
from concept checks and guided application to low-detail feedback and
localized clarification. Each feedback request creates a stateful
episode linked to versioned ERD states. In a deployment spanning three ERD environments and 383 feedback
episodes, 71.1\% of observed target-level changes fully or partially
incorporated the hidden diagnostic target, including many after
Stages~1--2. Qualitative analysis showed that staged disclosure
sometimes withheld inaccurate details, supported selective uptake, or
allowed later recovery, though some errors still shaped revisions.
Survey responses from a self-selected sample favored delayed disclosure
and student agency but noted indirectness and repetition.


\end{abstract}







\maketitle

\section{Introduction}
\label{sec:introduction}

Conceptual database design requires students to translate
natural-language requirements into entities, attributes, relationships,
and semantic constraints. This translation is rarely completed in one
step: students construct partially correct entity--relationship
diagrams (ERDs), reconsider individual modeling decisions, and revise
their artifacts as their interpretation of the requirements develops.
Timely formative feedback is therefore valuable, but providing
individualized guidance at the pace of revision remains difficult in
large courses
\cite{powerfeedback,kluger1996effects}. Automated feedback
\cite{keuning2018systematic,deorio2022automated} can expand access to
support, yet its educational value depends not only on whether a
diagnosis is correct, but also on when and how that diagnosis is
presented.

This distinction matters because feedback can fail in opposite ways.
A response that immediately reveals the correction may resolve the
current diagram while bypassing an opportunity for the student to
identify and apply the underlying concept. Feedback that remains too
indirect, however, may not provide enough information to support
revision. Adaptive scaffolding addresses this tension by increasing
support as needed
\cite{azevedo2004does,van2010scaffolding,pol2013scaffolding}, while
graduated prompting organizes that support from conceptual prompts to
task-specific cues and, when necessary, direct remediation
\cite{campione1987linking,ferrara1986children,resing2011dynamic}.
The appropriate progression nevertheless depends on the instructional
goal, assignment, and educator's feedback policy.

Large language models (LLMs) make such progression easier to implement
because they can interpret requirements, reason over structured
artifacts, and adapt explanations to an ongoing interaction. At the
same time, allowing the model to determine both the content and the
instructional strategy introduces risks of unsupported feedback,
over-assistance, premature solution disclosure, and limited educator
control
\cite{chu-etal-2025-llm,yang2025llm,yuan2025codae,
cohn2026theory,baek2026maskingagent}. These risks are particularly
relevant in ERD design, where knowing a concept does not ensure that a
student can apply it to a requirement or represent it correctly in a
diagram. A student may, for example, define total participation
correctly yet confuse it with an identifying relationship when revising
the artifact. One-shot feedback collapses conceptual review,
requirement interpretation, and diagram correction into a single
response, whereas unrestricted dialogue leaves their sequencing to the
LLM.

A recent requirement-grounded ERD feedback approach addresses the
evidence problem by combining structured student diagrams with
educator-authored requirements, rubrics, and seed questions
\cite{riazi&rooshenas25}. It identifies the requirements relevant to a
selected relationship and generates contextualized feedback without
requiring the student's ERD to match a single reference diagram.
Grounding determines what evidence supports the diagnosis, but it does
not determine how much of that diagnosis should be disclosed, what the
student should do before receiving a correction, or how later support
should depend on earlier responses.

We address this gap with an educator-guided architecture that places
artifact-grounded diagnosis inside an explicit, stateful pedagogical
workflow. The educator defines the stages, student actions, and
disclosure limits, while the LLM selects or generates content within
those constraints. We instantiate the architecture in an ERD editor
through a four-stage, learner-controlled workflow that progresses from
concept checking and guided application to low-detail feedback and
localized clarification. Feedback episodes are linked to versioned ERD
states, enabling analysis of how staged interactions relate to
subsequent revisions.

This work contributes an architecture that separates diagnosis,
pedagogical control, and student-facing generation; a reusable
execution model that combines structured artifact context,
educator-authored resources, and episode history; and a classroom
deployment examining workflow use, hidden-target incorporation,
diagnostic\hyp error behavior, and student perceptions. 
\section{Background and Related Work}
\label{sec:background}

\noindent\textbf{Automated Feedback and Intelligent Tutoring Systems.}
Intelligent tutoring systems (ITSs) commonly separate domain
representation, learner-state tracking, and pedagogical action
selection. Dialogue-based systems such as AutoTutor and its successors
demonstrate how this separation supports stepwise assistance across
domains
\cite{graesser1999autotutor,graesser2004autotutor,
rus2015deeptutor,rus2013recent,d2012gaze,olney2010tutorial,
graesser2003autotutor,sullins2015exploring,nye2014autotutor},
while constraint-based and Bayesian systems use structured knowledge
representations to diagnose student work and select feedback
\cite{mitrovic2007intelligent,bntutor}. Such systems provide substantial
pedagogical control but often require considerable domain-model
authoring and maintenance
\cite{paladines2020systematic,latif2026systematic,
zerkouk2025comprehensive}.

Database-design tutors such as KERMIT and EER-Tutor encode constraints
over entities, relationships, keys, cardinalities, and participation
and use interaction histories to guide subsequent support
\cite{mitrovic2006large,suraweera2002kermit,
weerasinghe2011facilitating}. Their precision, however, depends on
domain-specific authoring and often on canonical solution structures,
whereas valid ERDs may differ in naming, abstraction, and organization.
Prior LLM-based ERD feedback addresses this variation by evaluating
structured diagrams against educator-authored requirements, rubrics,
and questions rather than a single reference solution
\cite{riazi&rooshenas25}. The present work retains this
requirement-grounded diagnosis but focuses on how it is delivered
through a controlled instructional workflow.

\noindent\textbf{Graduated Prompting and Adaptive Scaffolding.}
Adaptive scaffolding varies the amount and specificity of assistance
according to learner performance
\cite{azevedo2004does,van2010scaffolding,pol2013scaffolding}.
Graduated prompting makes this progression explicit by moving from
general conceptual prompts to task-specific cues and, when necessary,
direct remediation
\cite{campione1987linking,ferrara1986children,
resing2011dynamic}. The aim is to provide the least explicit support
under which the learner can continue.

Our workflow applies this structure to artifact-centered feedback. A
hidden diagnosis identifies the targeted ERD issue, while successive
stages control both the instructional activity and the amount of
diagnostic information that may be disclosed. Unlike procedures that
advance automatically after an unsuccessful response, progression is
learner-controlled: after each interaction, students may revise the ERD
or request more explicit assistance. Each stage can therefore serve as
a stopping point rather than a mandatory step.

\noindent\textbf{LLM-Based Educational Feedback and Pedagogical Agents.}
LLMs have been used for instructional-material generation
\cite{leiker2023prototyping}, formative feedback
\cite{feedback_chatgpt,jamal2024enhancing,
letteri2025enhancing}, programming exercises
\cite{sarsa2022automatic}, guided questioning
\cite{ochieng2023large,cui2025loom}, and assessment support
\cite{moore2022assessing,gonnermann2025facet}. Synthetic Educational
Feedback Loops~\cite{zhang2025sefl} similarly generate teacher--student exchanges for
training feedback models, but do not execute individualized,
multi-stage tutoring workflows.

Recent pedagogical agents represent instructional state explicitly.
Cohn et al. (2026) combine learning theories with assessment and adaptive
decision modules, using dialogue evidence to guide response selection
\cite{cohn2026theory}. Yang et al. (2025) model discussion stages,
learning issues, and intervention timing for collaborative feedback
\cite{yang2025llm}. Our system instead organizes state around one
diagnosed issue in a versioned student artifact. Revising the ERD ends
the current episode and causes a later request to be diagnosed as a new episode,
rather than preserving the previous error state. An educator-defined
controller determines the active interaction, available context, and
disclosure level before generation.

Model adaptation offers another means of instructional alignment.
CoDAE\cite{yuan2025codae} fine-tunes LLMs on augmented educational dialogues to reduce
over-compliance, handle uncertainty, and resist premature answer
requests. Our approach leaves the model unchanged and
places generation within an external workflow policy. These strategies
are complementary: model adaptation influences general behavior,
whereas workflow control specifies what assistance is permitted in a
particular episode.

\noindent\textbf{Reliability, Disclosure, and Educator Governance.}
General-purpose LLMs may omit errors, produce unsupported explanations,
or reveal more of a solution than intended
\cite{sessler2025towards,renze2024self,milano2023large}.
MaskingAgent~\cite{baek2026maskingagent} addresses disclosure after generation by detecting and
removing essential parts of an overly complete programming solution. Our system instead limits what diagnostic information the LLM can use
and what type of support it may generate at each stage. Because the LLM
may still produce content that exceeds these limits, the system logs
the active stage, input context, student response, generated feedback,
and disclosed information for later review. Educator governance is expressed through requirements, rubrics,
concept-aligned questions and resources, workflow definitions, and
feedback-credit policies. The agent does not modify the ERD, and
students may return to the editor after any interaction. The design
therefore preserves student authorship while making the timing,
specificity, and extent of automated assistance explicit and
configurable.
\begin{figure*}[t]
    \centering
    \includegraphics[width=\linewidth]{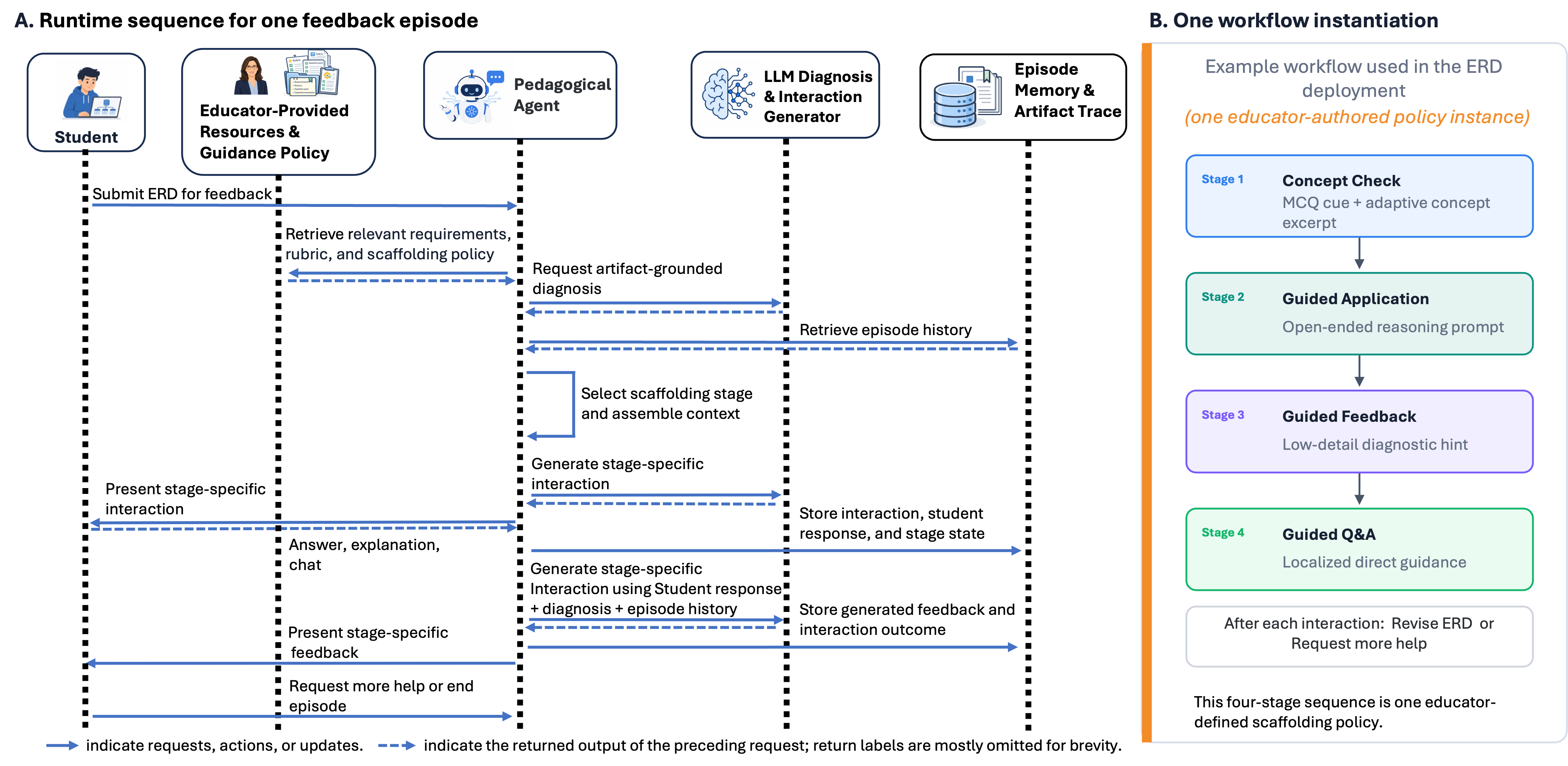}
    \caption{(A) Runtime sequence linking artifact-grounded diagnosis,
    workflow control, episode memory, and student interaction.
    (B) Four-stage workflow used in the classroom deployment.}
    \label{fig:architecture}
\end{figure*}

\section{System Design and Implementation}
\label{sec:system}

The system adds an educator-defined workflow layer to a
requirement-grounded ERD feedback pipeline. As shown in
Figure~\ref{fig:architecture}, each request creates an episode linking
the current artifact, a hidden diagnosis, the active pedagogical stage,
and the resulting interaction. The controller selects the stage and
permitted disclosure, while the presented content may be
educator-authored, template-instantiated, or LLM-generated.

\subsection{Artifact Representation and Diagnostic Context}
\label{sec:context}

Each assignment provides student-facing requirements and
educator-authored rubrics. Question pools, conceptual explanations, and
workflow stages are defined at the learning-objective level and can be
reused across assignments that address the same objective. Educators
therefore do not need to create a separate question pool for every
assignment.

The editor parses each ERD into a structured representation of its
entities, attributes, keys, relationships, participating entities,
cardinalities, participation constraints, and entity types. This
representation is used both to render the diagram and to analyze it
independently of its layout or the exact names chosen by the student.

When a student requests feedback for a selected relationship, the
system retrieves the relevant requirement and rubric and keeps only the
selected relationship and its participating entities. The resulting
\textit{context} contains the hidden diagnosis, whether an error was
detected, the identified error categories, the relevant ERD fragment,
and the associated requirement and rubric. This information remains
internal, and the controller exposes only the parts permitted by the
active stage.

The diagnostic pipeline is adopted from a prior requirement-grounded
ERD feedback system \cite{riazi&rooshenas25} and is not re-evaluated
here. The current implementation retains the same use of structured
artifacts, requirements, and rubrics but represents constraints more
explicitly. For example, total participation is stored in a dedicated
key--value field rather than only as a label in a general type list.
This change was intended to reduce ambiguity for the LLM; we do not
claim a measured improvement over the earlier representation. The
present work focuses on how the resulting diagnosis is delivered
through an educator-defined workflow.

\vspace{-0.01in}
\subsection{Educator-Defined Workflow Execution}
\label{sec:workflow}

An educator defines the sequence of stages available to students, such
as concept checks, application questions, hints, explanations, revision
checkpoints, and guided dialogue. For each stage, the educator specifies
what the student should do, what information may be shown, what response
should be recorded, and what actions may follow.

At runtime, the controller selects the active stage and assembles the
\textit{context} available to the LLM from the ERD diagnosis,
educator-provided resources, and retrieved episode history. The
controller determines what information may be disclosed, while the LLM
selects, adapts, or generates content within those limits.

The ERD diagnosis and episode-history components are reusable across
different workflows. An instructor can therefore skip or repeat a
stage, replace one stage with another type of support, or branch based
on a student's response without rebuilding the diagnostic or memory
components.

Staged disclosure can also reduce students' exposure to some diagnostic
errors. Early stages rely mainly on educator-authored or template-based
conceptual content, so an incorrect localized correction may remain
hidden. This protection is limited: if the system selects the wrong
relationship or concept, the resulting questions may still be
irrelevant or misleading. The no-error route is also unbuffered because
a false negative is communicated as confirming feedback.

\begin{figure}[t]
    \centering
    \includegraphics[width=\columnwidth]{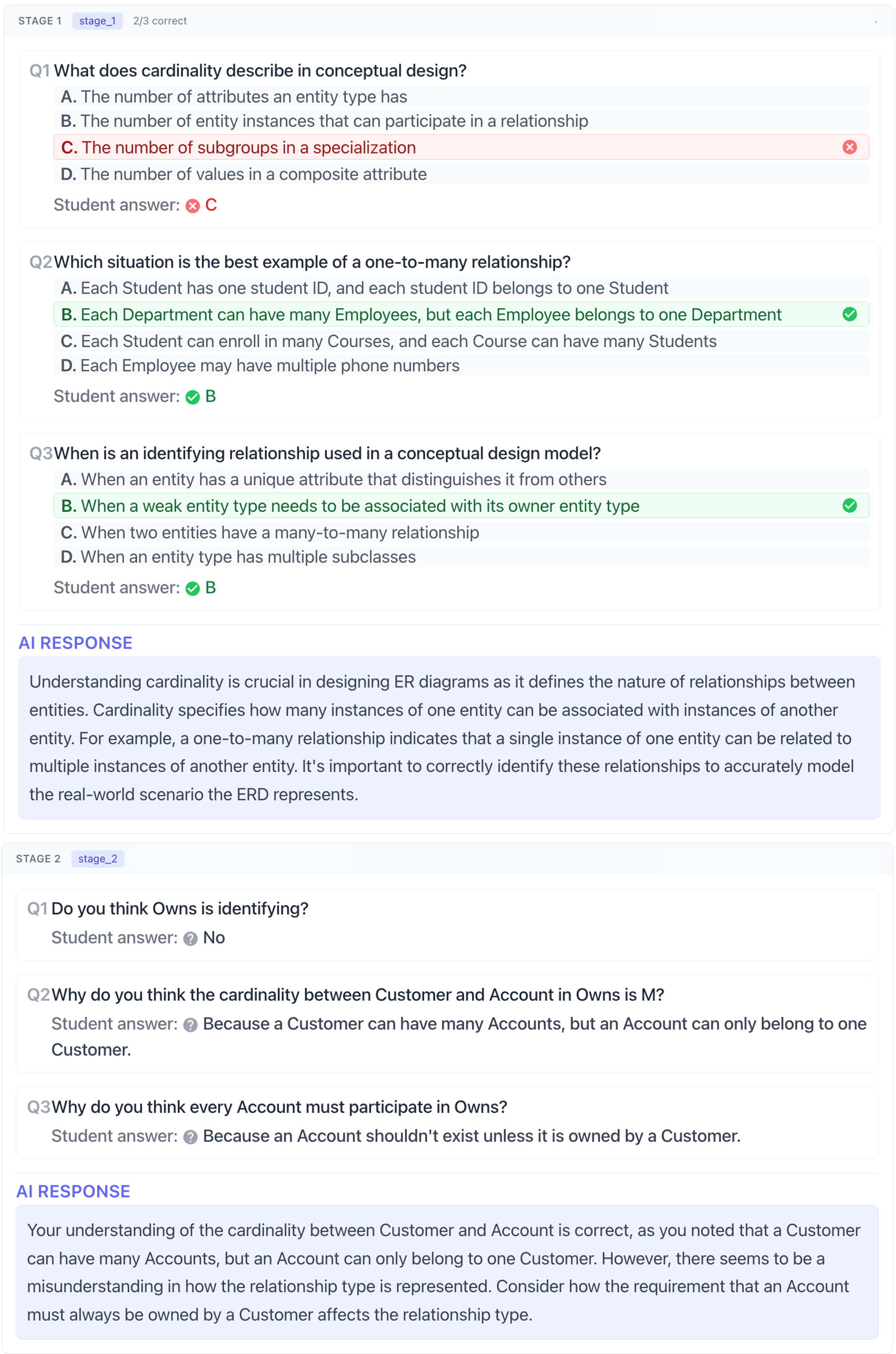}
    \caption{Student interaction log for Stages~1 and~2 with system feedback.}
    \label{fig:student-interaction}
\end{figure}

\subsection{Four-Stage Workflow}
\label{sec:scaffold}

The deployed workflow follows graduated prompting under learner control
\cite{campione1987linking,ferrara1986children,resing2011dynamic}.
Each accessed stage consumes one feedback credit. After every
interaction, the student may revise the ERD or request the next, more
explicit form of assistance.

\noindent\textbf{Stage 1: Concept Check.} Stage~1 provides the LLM with the complete educator-authored question
pool defined for that stage. Using the context assembled by the
controller, the LLM selects up to three relevant questions but does not
generate or modify their content. The questions, answer keys, and
conceptual explanations are pre-authored and do not name the targeted
ERD component, quote its requirement, or reveal a correction.

\noindent\textbf{Stage 2: Guided Application.}
Stage~2 uses educator-authored question templates whose context-specific
slots are filled from the selected ERD component. For example, the
template ``Do you think \texttt{[Relationship]} is identifying?'' becomes a question about the relationship under review. Each
template includes a reference rubric describing the relevant reasoning,
such as whether one entity is weak, whether it requires the owner's key
for identification, and whether it can be uniquely identified by its
own attributes. The LLM evaluates the student's written response
against this rubric and generates a brief closing statement, while the
localized artifact correction remains hidden.

\noindent\textbf{Stage 3: Guided Feedback.}
The controller assembles the context permitted for Stage~3 and requests
a concise, artifact-grounded response from the LLM. For error-positive
cases, the response identifies the general issue and poses a focused
revision question while withholding the localized correction. If no
error is diagnosed, the controller routes directly to Stage~3 and
requests confirming feedback. Because this route communicates the
no-error judgment without earlier scaffolding, false negatives are not
buffered by the preceding stages.

\noindent\textbf{Stage 4: Guided Q\&A.}
The final stage supports multi-turn, requirement\hyp grounded
clarification. For each turn, the controller assembles the permitted
artifact context and episode history, and the LLM generates a response
to the student's question. The response may name the affected
components and recommend a localized modeling action, but the agent
does not generate or modify the complete ERD.

Figure~\ref{fig:student-interaction} shows the progression from concept checking to assignment\hyp specific reasoning while keeping the diagnosis hidden.

\subsection{Episode Memory and Artifact Traces}
\label{sec:memory}

Each request creates a \textit{session\_history} containing the stages
shown, student answers and written responses, generated feedback,
follow-up questions, credit use, and selected next actions. The
\textit{context} grounds the interaction in the current ERD
and requirements, while the session history allows later stages to
condition on what has already occurred. These records coordinate the
current episode and are not treated as a persistent model of the
student's knowledge.

Returning to the editor ends the episode. A later request reprocesses
the current ERD and creates a new episode, even when the same
relationship is selected, so the previous diagnosis is not assumed to
remain valid after revision. Episodes are linked to saved artifact
states through the student, environment, selected relationship,
feedback identifier, and ERD version. The resulting
artifact--interaction--revision traces contain the initiating ERD,
hidden context, stages accessed, student responses, generated
interactions, and next saved ERD, supporting analysis of progression,
disclosure, repetition, and subsequent target-level changes. 
\section{Classroom Deployment}
\label{sec:deployment}

The system was deployed in an online, mixed-level Database Systems
course in Summer 2026. The course enrolled 63 students, predominantly
undergraduates. Forty-eight students completed an individual homework
containing three ERD environments assigned together for one week. Use
of the editor was required, but AI feedback was optional and ungraded;
teaching assistants independently graded the submitted ERDs.

Students received written instructions and a demonstration of all four
stages. Each student received 15 feedback credits per environment, and
each accessed stage consumed one credit. No student requested
additional credits. GPT-4o was used for the LLM-mediated diagnosis and
stage-generation steps.

A feedback episode began when a student selected a relationship and
requested feedback. Returning to the editor and requesting feedback
again created a new episode, even for the same relationship, causing
the current ERD to be processed again. Logs were restricted to the
homework period, with instructor, teaching-assistant, and test accounts
removed.

\subsection{Trajectory Analysis}

For each episode, we compared the ERD at feedback time with the first
subsequently stored ERD, when available. 
The comparison focused on the selected relationship and its local
properties, including participating entities, cardinality,
participation constraints, identifying status, and applicable
specialization/union structures.
GPT-4o received the target-level difference together with the original hidden
diagnosis, error categories, relevant requirement, and targeted ERD
fragment, and classified the subsequent state as fully, partially, or
not incorporating the hidden diagnostic target.

This use of the LLM differs from using an LLM judge to determine
whether generated feedback or a revised ERD is correct. The classifier
was given the hidden target as the reference and performed the narrower
task of determining whether the observed revision corresponded to that
target. Its verdict therefore measures behavioral alignment with the
hidden diagnosis, not the correctness of the diagnosis, the quality of
the revision, student learning, or fidelity to the exact
student-visible scaffold. Because some diagnostic details may have
been withheld at the stage reached, partial incorporation also does not
necessarily mean that a student ignored an explicit instruction.

Episodes with no target-level update were separated according to
whether a later ERD had been stored, and no-error diagnoses were
analyzed separately. We manually inspected 20 randomly selected
trajectories by comparing the feedback-time ERD, next saved ERD, hidden
diagnosis, student interactions, scaffolded feedback, and assigned
verdict. We found no disagreements in this sample. This check supports
the consistency of the incorporation classification relative to the
hidden target; it does not validate the correctness of the underlying
diagnosis.

\begin{table}[t]
\caption{Deployment and revision outcomes across three ERD
environments. Each episode concerns one relationship.}
\label{tab:deployment-outcomes}
\vspace{-0.1in}
\centering
\small
\setlength{\tabcolsep}{1.1pt}
\renewcommand{\arraystretch}{1.06}
\begin{tabular}{@{}lrrrr@{}}
\toprule
\textbf{Measure} &
\textbf{Env.\ 1} &
\textbf{Env.\ 2} &
\textbf{Env.\ 3} &
\textbf{All} \\
\midrule
\multicolumn{5}{@{}l}{\textit{Deployment}} \\
Feedback users
    & 17 & 24 & 32 & -- \\
Episodes
    & 100 & 149 & 134 & 383 \\
Median/user
    & 5 & 4 & 3 & -- \\
Later ERD stored
    & 58 (58.0\%) & 84 (56.4\%) & 44 (32.8\%) & 186 (48.6\%) \\
\midrule
\multicolumn{5}{@{}l}{\textit{Hidden-target incorporation}} \\
Fully incorporated
    & 18 (18.0\%) & 26 (17.4\%) & 8 (6.0\%) & 52 (13.6\%) \\
Partly incorporated
    & 8 (8.0\%) & 20 (13.4\%) & 11 (8.2\%) & 39 (10.2\%) \\
Not incorporated
    & 6 (6.0\%) & 10 (6.7\%) & 21 (15.7\%) & 37 (9.7\%) \\
No target update, later ERD
    & 7 (7.0\%) & 11 (7.4\%) & 1 (0.7\%) & 19 (5.0\%) \\
No target update, no later ERD
    & 13 (13.0\%) & 32 (21.5\%) & 58 (43.3\%) & 103 (26.9\%) \\
No error diagnosed
    & 48 (48.0\%) & 50 (33.6\%) & 35 (26.1\%) & 133 (34.7\%) \\
\bottomrule
\end{tabular}

\vspace{2pt}
{\footnotesize\raggedright
User counts are not pooled because students may appear in multiple
environments. Incorporation is evaluated against the complete hidden
diagnostic target and does not imply correctness.\par}
\end{table}

\subsection{Trajectory Results}

Table~\ref{tab:deployment-outcomes} shows that students used the agent
repeatedly across all three environments, but observable revision was
uneven. A later ERD was available for 186 of 383 episodes (48.6\%),
indicating that many requests were followed either by no saved revision
or by work that could not be linked to a subsequent artifact state.

Among the 250 episodes with an identified error, 128 (51.2\%) included
a change to the targeted relationship or participation constraint.
Ninety-one of these changes fully or partially incorporated the hidden
diagnostic target. Thus, incorporation occurred in 36.4\% of all
error-positive episodes and in 71.1\% of those with an observed
target-level change. The latter rate should therefore be interpreted as
conditional on the student making a relevant revision, not as an
overall success or correctness rate.

Students generally used the workflow as an early-stage scaffold rather
than progressing to its most explicit forms. Most responding episodes
ended after Stage~1 or Stage~2, while only 12 error-positive episodes
reached Stage~4. Among the 109 target-level changes following these
early exits, 80 (73.4\%) incorporated at least part of the hidden target.
Because the explicit correction had not yet been disclosed, these
episodes suggest that some students could revise after conceptual or
requirement-level support alone. However, the incorporation measure
does not establish that the hidden diagnosis was correct or that every
target detail had been communicated to the student. Thirty-five
episodes also ended after the concept check was displayed but before
the student responded, showing that early presentation did not always
lead to continued interaction.

The third environment showed a different pattern, with fewer later
saved ERDs, more episodes without a target-level revision, and lower
early-stage incorporation than the other two environments. Its
requirements contained several closely related relationships that
required distinctions among direction, cardinality, and participation.
This may have increased the difficulty of translating conceptual
guidance into a diagram, although the environments were not designed as
comparison conditions. The differences may therefore reflect task
structure, submitted artifacts, timing, or student behavior rather than
a systematic limitation of the workflow.

\subsection{Qualitative Analysis of Diagnostic Errors}

During manual inspection of randomly sampled trajectories, we examined
the ERD at feedback time, the next saved ERD, the hidden diagnosis, the
student's stage interactions, and the scaffolded feedback shown to the
student. The cases below illustrate how these elements can diverge;
they do not estimate the prevalence of each pattern.

\noindent \textbf{Incorrect detail not exposed.}
In some Stage~1 trajectories, an inaccurate hidden diagnosis was used
to select a related conceptual topic rather than being shown as
detailed diagram feedback. 
For example, one hidden diagnosis incorrectly
stated that total participation was not required for a strong entity in
a non-identifying relationship. The student answered the
total-participation concept question incorrectly, and the visible
Stage~1 scaffold provided an accurate educator-authored explanation
without exposing the erroneous claim. The next saved ERD did not
include the participation change incorrectly recommended by the hidden
diagnosis.

\noindent\textbf{Contradictory diagnoses not directly disclosed.}
In one sequence, the first hidden diagnosis incorrectly treated total
participation as evidence that a relationship should be identifying,
and the student subsequently made that change. The revised ERD no
longer contained the pattern that appeared to trigger the earlier
incorrect inference, so a new request produced the opposite, correct
diagnosis. Because neither diagnosis was presented verbatim, the
student did not receive two explicit and contradictory corrections.
The low-detail interaction instead left room for another feedback
request, after which the student corrected the relationship.

\noindent\textbf{Selective incorporation.}
Some hidden diagnoses combined correct and incorrect recommendations.
In one trajectory, the student implemented only the valid
participation change and did not adopt an incorrect recommendation to
make the relationship identifying. The classifier labeled the episode
partially incorporated because the revised ERD matched only part of the
complete hidden target. In other partially incorporated cases, the
unapplied portion was not clearly stated in the student-visible
scaffold. Partial incorporation therefore does not necessarily mean
that a student disregarded explicit feedback.

\noindent\textbf{Incorrect target incorporated.}
In other trajectories, an inaccurate claim entered the
student-visible scaffold and the student revised the ERD accordingly.
The classifier labeled such a revision fully incorporated because it
matched the hidden diagnostic target, even though the target itself was
incorrect. Incorporation therefore measures correspondence with the
hidden diagnosis rather than correctness of the resulting ERD.

\noindent\textbf{Correct diagnosis without incorporation.}
The reverse pattern also occurred. In one trajectory, the diagnosis
and progressively more explicit feedback correctly identified a
specialization-related modeling issue. The student reached all four
stages and expressed relevant understanding in written responses, but
the next ERD changed nearby cardinalities rather than the relationship
participants. Correct and increasingly explicit feedback therefore did
not guarantee that the student could represent the concept in the
artifact.

These cases show why the hidden diagnosis, student-visible scaffold,
and subsequent revision must be considered separately. An incorrect
diagnostic detail may remain hidden, affect the scaffold indirectly, or
be explicitly incorporated by the student. Conversely, accurate
conceptual feedback may not lead to the intended artifact change.

\subsection{Student Survey}

An optional, ungraded survey was distributed to all 63 enrolled
students one week after the homework deadline and after grades were
released. Thirty-one students responded; one was excluded after
reporting that they had not used the agent, leaving at most 30 system
users (47.6\% of enrollment). Because participation was voluntary, the
responses may reflect self-selection. Identifying information was
removed before analysis, and item-level sample sizes vary because some
respondents skipped questions. The study was reviewed and approved as
exempt by the institutional IRB.

\begin{table}[t]
\caption{Student perceptions of the staged feedback workflow.}
\label{tab:survey-results}
\vspace{-0.1in}
\centering
\small
\setlength{\tabcolsep}{1.8pt}
\renewcommand{\arraystretch}{1.08}
\begin{tabular}{@{}p{0.75\columnwidth}rrrr@{}}
\toprule
\textbf{Survey item} &
\textbf{$n$} &
\textbf{A} &
\textbf{N} &
\textbf{D} \\
\midrule
Understood the conceptual-design topic
    & 29 & 17 & 8 & 4 \\
Early stages allowed time to think before correction
    & 29 & 15 & 8 & 6 \\
Could make my own design decision
    & 28 & 18 & 9 & 1 \\
Trusted the feedback enough to consider it
    & 28 & 16 & 6 & 6 \\
Would use staged feedback in future assignments
    & 28 & 16 & 7 & 5 \\
System revealed too much information too early
    & 28 & 0 & 7 & 21 \\
System remained too indirect when I needed an answer
    & 28 & 11 & 12 & 5 \\
System repeated questions or explanations
    & 28 & 12 & 10 & 6 \\
\bottomrule
\end{tabular}

\vspace{2pt}
{\footnotesize\raggedright
A = agree or strongly agree; N = neutral; D = disagree or strongly
disagree. Item-level sample sizes vary because some respondents skipped
questions.\par}
\end{table}

Table~\ref{tab:survey-results} presents selected perceptions from this
self-selected sample. Eighteen of 28 respondents (64.3\%) agreed that
they could make their own design decisions, and 21 of 28 (75.0\%)
disagreed that the system revealed too much information too early.
More than half reported understanding the topic being addressed
(17/29, 58.6\%), trusting the feedback enough to consider it
(16/28, 57.1\%), and willingness to use staged feedback in future
assignments (16/28, 57.1\%). Fifteen of 29 (51.7\%) agreed that early
stages allowed time to think before receiving a correction.

The responses also identified limitations. Eleven of 28 respondents
(39.3\%) found the system too indirect when they needed a more explicit
answer, and 12 of 28 (42.9\%) reported repeated questions or
explanations. Open-ended responses similarly described conceptual
questions as useful for reconsidering cardinality and participation,
while requesting more localized guidance, smoother transitions, and
less repetition.

\section{Conclusion}
\label{sec:conclusion}

This work demonstrates how an LLM-based feedback system can support
artifact-centered revisions without making the language model solely
responsible for the instructional strategy. By placing diagnosis,
workflow progression, disclosure level, and interaction memory within
an educator-guided architecture, the system supports feedback that can
begin with conceptual reflection and become more explicit only when
students seek additional assistance.


The classroom deployment shows that students sometimes reconsidered
and revised local design decisions after staged support while retaining
responsibility for the final artifact.
It also highlights the importance of making transitions
between stages responsive and avoiding unnecessary repetition. Future
work will refine these transition mechanisms, improve workflow
authoring support for educators, and examine how the architecture can
be adapted to other structured design tasks and instructional
contexts.

\section*{Ethics and Privacy Statement}

The study was reviewed by the institutional IRB and determined to be exempt. Survey responses
and interaction logs were de-identified, and the optional, ungraded
survey did not affect students' grades. The system was used only for
formative support, with teaching assistants independently grading
submissions; future deployments should continue monitoring feedback
reliability, accessibility, and differential use.
\clearpage

\bibliographystyle{ACM-Reference-Format}
\bibliography{references}


\end{document}